\documentclass{article}

\usepackage{arxiv}

\usepackage[utf8]{inputenc} 
\usepackage[T1]{fontenc}    
\usepackage{hyperref}       
\usepackage{url}            
\usepackage{booktabs}       
\usepackage{amsfonts}       
\usepackage{nicefrac}       
\usepackage{microtype}      
\usepackage{lipsum}		
\usepackage{graphicx}
\usepackage{natbib}
\usepackage{doi}
\usepackage{amsthm}
\usepackage{xspace}
\usepackage{float}
\usepackage{wrapfig}
\usepackage{verbatim}

\usepackage{pifont}
\newcommand{\yes}{\ding{52}}
\newcommand{\no}{\ding{56}}
\newcommand{\yn}{partial}

\newtheorem{definition}{Definition}

\def\newxai{evaluative AI\xspace}
\def\Newxai{Evaluative AI\xspace}
\def\dst{DST\xspace}
\def\dsts{DSTs\xspace}

\title{Explainable AI is Dead, Long Live Explainable AI!\\
\Large{Hypothesis-driven decision support}}
\renewcommand{\shorttitle}{Explainable AI is Dead, Long Live Explainable AI!}

\author{Tim Miller\\
School of Computing and Information Systems\\The University of Melbourne, Melbourne, Australia\\\url{tmiller@unimelb.edu.au}}

\begin{document}
\maketitle

\begin{abstract}
In this paper, we argue for a paradigm shift from the current model of explainable artificial intelligence (XAI), which  may be counter-productive to better human decision making. In early decision support systems, we assumed that we could give people recommendations and that they would consider them, and then follow them when required. However, research found that people often ignore recommendations because they do not trust them; or perhaps even worse, people follow them blindly, even when the recommendations are wrong. Explainable artificial intelligence mitigates this by helping people to understand how and why models give certain recommendations. However, recent research shows that people do not always engage with explainability tools enough to help improve decision making. The assumption that people will engage with recommendations and explanations has proven to be unfounded. We argue  this is because we have failed to account for two things. First, recommendations (and their explanations) take control from human decision makers, limiting their agency. Second, giving recommendations and explanations does not align with the cognitive processes employed by people making decisions. This position paper proposes a new conceptual framework called \textbf{\Newxai} for explainable decision support. This is a machine-in-the-loop paradigm in which decision support tools provide evidence for and against decisions made by people, rather than provide recommendations to accept or reject. We argue that this mitigates issues of over- and under-reliance on decision support tools, and better leverages human expertise in decision making.
\end{abstract}




\section{Introduction}

Imagine you have two friends, Bluster and Prudence. Whenever you have a difficult decision to make, you can approach them for help. Both have shown excellent judgement on complex decisions in the past. Bluster always tells you what they think is the right decision, even if they are not confident, and then tells you why they think that. Would that be helpful? If your answer and reasons were the same as Bluster's, it would give you confidence. Bluster could change your mind to an answer that you were happier with or that resulted in  better outcomes. Prudence, in contrast, hardly ever gives their opinion, especially when not confident. Prudence instead asks you what you are proposing and then provides feedback: evidence for and against your proposed decision. If you propose alternatives, Prudence continues to provide  feedback on these, giving feedback until you reach a decision. But Prudence never gives you an answer. Would this be more useful than Bluster's approach? Prudence helps you to form a decision, and provides feedback on your own options, rather than justifying their own opinion. This would help you to question your decisions, and would give you control over which options you receive feedback for.

Reader, which would you prefer? A survey conducted during a recent talk showed that just three out of over 100 people preferred Bluster; the remaining 100+ people preferring Prudence. Prudence helps us find strengths and weaknesses in our thinking and gives us control over which options we discuss with them. The ability to assess the strengths and weaknesses of judgements and decisions is key to expert decision making \citep{Klein2017-uw,Graber2012-uy,Lambe2016-oz}.

Despite this preference, the current model of (explainable) AI-assisted decision support gives us Bluster instead of Prudence, right? AI-assisted decision tools are designed to tell the user what it thinks the best answer is (e.g.\ a recommendation), and explain why that is considered the best answer, even when they are not confident that their recommendation is correct. What if the user disagrees with this answer and doesn't find the reasons convincing? The machine offers little else. A counterfactual explanation \citep{Miller2021-ix} allows us to ask why another option is not the best answer; but does not provide us with reasons this alternative may well be the \emph{right} answer, or even a good answer. In short, current approaches to explainable AI, which we call \textbf{recommendation-driven} decision support, do little to help us critique its answers or our own ideas. 

\begin{wrapfigure}{R}{0.5\linewidth}
\centering
\includegraphics[scale=0.5]{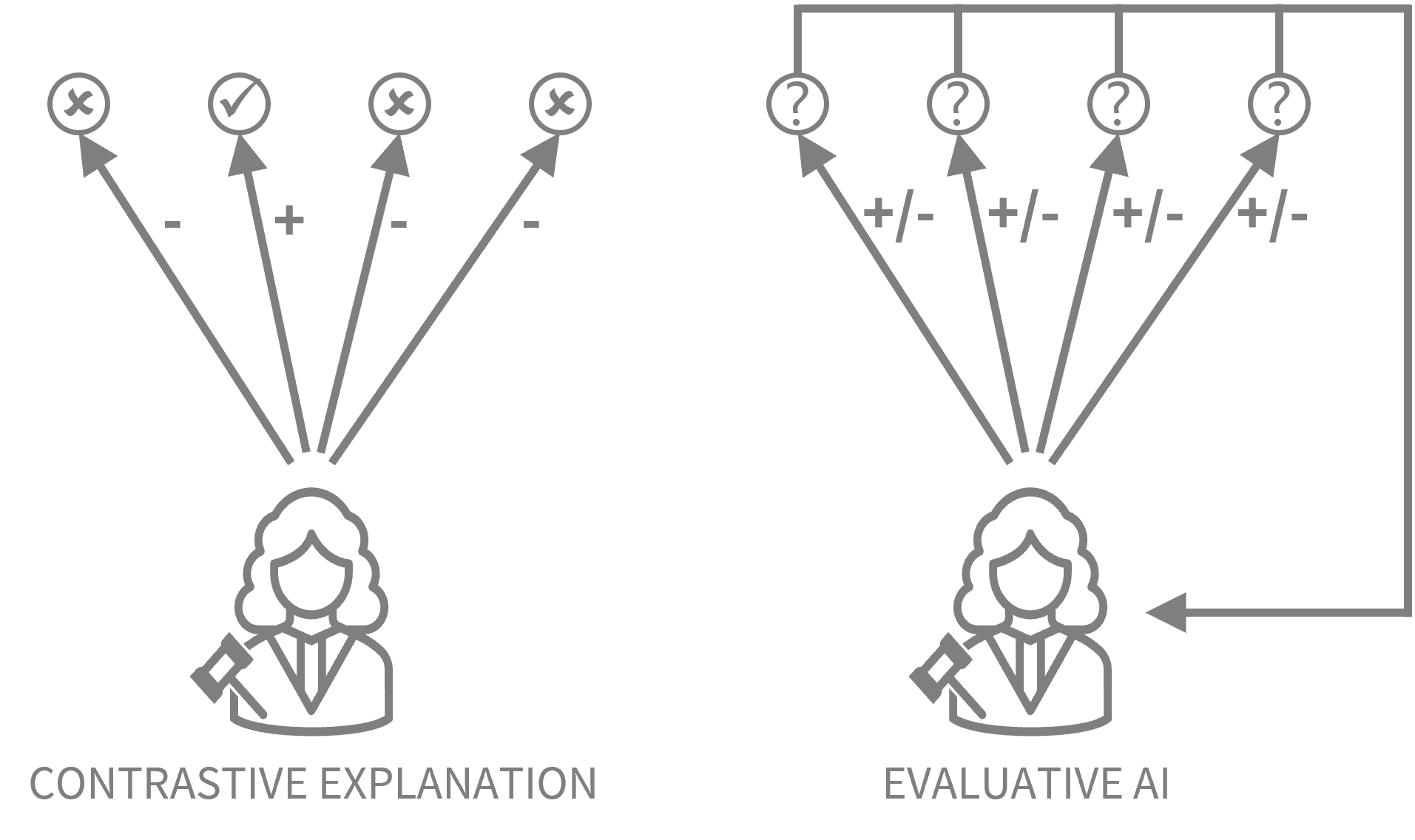}
\caption{Contrastive explanation vs \Newxai. Contrastive explanation paradigms are best described as a process of persuading us to accept a machine recommendation. They only provide evidence that supports the recommendation and refutes all others. \Newxai proposes not necessarily giving recommendations from machines, but instead providing evidence for/against each option.}
\label{fig:contrastive-vs-evaluative}
\end{wrapfigure}

The Bluster model of recommendation-driven decision support leads to two related problems. First, it is hard for people to know how much to trust recommendations \citep{Jacovi2021-nw,Hoffman2017-fn,Lee2004-fz,Sivaraman2023-ea}.  AI tools are not correct all of the time, so we should be sceptical at least some of the time. However, people find it difficult to correctly calibrate their trust in a decision aid. Research shows that incorrect trust calibration people tend to either \emph{under-rely} on tools, meaning they have no effect on decision making, or they \emph{over-rely} on tools \citep{Bucinca2021-lq,Gajos2022-is,Sivaraman2023-ea}, likely because providing recommendations makes them \textbf{fixate} on that recommendation. Both under- and over-reliance have negative consequences \citep{Parasuraman1997-th}.  Second, Bluster reduces our locus of control \citep{Shneiderman2016-bn}, because we cannot control which options we received feedback for.

This paper argues for a paradigm shift: that the decision support tools (\dsts) should be more like Prudence, which we call \textbf{hypothesis-driven} decision support, and less like Bluster. We propose a new conceptual framework for explainable decision support that has two key properties not present in the  current paradigm. First, \newxai tools do not provide recommendations. They either allow the decision maker to determine which options are best or help them  to narrow down to a manageable set of options \citep{Rudolph2003-tu}. This helps to mitigate fixation caused by things like automation bias. Second, instead of justifying AI recommendations, \dsts generate and present evidence to support or refute human judgements, and explain trade-offs between any set of options, not just the machine recommendation. This helps with trust calibration because the machine does not give recommendations, as well as over/under-reliance because there is no recommendation to follow.
We argue that the reason this Prudence-like approach is more effective for decision support because it \textbf{aligns with the cognitive decision-making process that people use when making judgements and decisions} \citep{Klein2007-bs,Hoffman2022-en,Peirce2009-zg}. We call this paradigm \textbf{\newxai}.

Figure~\ref{fig:contrastive-vs-evaluative} shows the difference between \newxai and perhaps the most common form of explainability: contrastive explanation. Contrastive explanation is like Bluster --- it gives us an answer and justifies it, telling us why it is correct and why other options are not. \Newxai is like Prudence --- it helps us critique our own ideas. This provides a better feedback loop -- if the evidence helps use eliminate our preferred hypothesis, we start to explore others.

\Newxai is not intended to be used in all scenarios. It is more suitable for medium- and high-stakes decisions when human decision makers are ultimately accountable, and low frequency decision making where the decision maker has time to explore options.

We conclude that for decision making, the recommendation-driven paradigm of explainable AI is `dead' (for some situations), but that the \newxai paradigm is still a form of explainable AI, and many of the current paradigm will play a part in this new conceptualisation; so, long live explainable AI!

Section~\ref{sec:background} reviews related work on cognitive decision making, what makes a good decisions, and the main modes of explainable AI. Section~\ref{sec:xai-as-decision-support} evaluates how current decision support models align with human decision making processes, with a focus on explainable and interpretable AI. Section~\ref{sec:newxai} presents \newxai, a new conceptual framework for human-centred decision making, and argues that this framework aligns better the cognitive processes humans use for decision making. It also presents a high-level research agenda for \newxai. Section~\ref{sec:conc} concludes the paper.

\section{Background and related work}
\label{sec:background}


\subsection{Decision making and decision support}

Before we consider what makes good decision support, we consider what good decision making is.
Table~\ref{table:deciding} outlines 10 \emph{cardinal decision issues}, defined by \citet{Yates2012-ax}. These include issues such as deriving options and making judgements, but also additional factors such as exploring consequences of actions, and who will be part of deciding. The issues in italics are the aspects that we believe to be of most importance to explainable decision support. While the first three and the final issue can be supported by tools, these are less relevant for AI research. 

\begin{table}[!th]
\centering
\caption{The 10 `cardinal decision issues' outlined by \citet{Yates2012-ax}. The issues in italics represent those that are of most interest to explainable AI.}
\label{table:deciding}
\begin{tabular}{ll}
\toprule
\textbf{Cardinal issue} & \textbf{Definition}\\
\midrule
 Need          & Why do we need to make a decision at all?\\[1mm]
 Mode          & Who will decide and how will they do it?\\[1mm]
 Investment    & What kinds of amounts of resources will be invested in the process?\\[1mm]
 \emph{Options}       & What are the different actions we could take to solve the need?\\[1mm]
 \emph{Possibilities} & What outcomes could happen if for each action, if it were taken?\\[1mm]
 \emph{Judgement}     & Which of the outcomes would happen if we took the action?\\[1mm]
 \emph{Value}         & How much would any stakeholder care (positively or negatively) if this outcome happened?\\[1mm]
 \emph{Trade-offs}    & How do we trade-off the outcomes to settle on an action?\\[1mm]
 \emph{Acceptability} & How can be get other stakeholders to accept our decision?\\[1mm]
 Implementation & Now we have decided, how can we action the decision?\\
 \bottomrule
\end{tabular}
\end{table}

We define a \textbf{decision support tool} (\dst) as any system that supports the process of deciding. But what is a \emph{good} \dst? In Table~\ref{tab:good-decision-support}, we propose six criteria for good decision support. The first five are based on the 10 cardinal issues from Table~\ref{table:deciding}
\citep{Hoffman2005-bk,Yates2003-lj}. The sixth is \emph{understandable}, which simply means that a good decision support tool helps people understand how and why it works, and where it fails. This is important to calibrate trust and find mistakes. The broader range of research into explainability, interpretability, and transparency focuses on this. 

\begin{table}[!ht]
\centering
\caption{Criteria for good decision support, based on the 10 `cardinal decision issues' \citep{Yates2012-ax}}
\label{tab:good-decision-support}
\begin{tabular}{ll}
   \toprule
    \textbf{Criteria} & \textbf{Definition}\\
    \midrule
    {Options} & Help to identify options, as well as help to narrow down the list of feasible or realistic options\\[1mm]
    {Possibilities} & Help to to identify possible outcomes for each of the identified options\\[1mm]
    {Judgement} & Help to judge which outcomes are most likely\\[1mm]
    {Value} & Help to identify the positive and negative impacts on stakeholders for each of the identified options\\[1mm]
    {Trade-offs} & Help to make trade-offs on the above criteria for each options\\[1mm]
    {Understandable} &  Help to understand how and why the tools works as it does, and when it fails\\
    \bottomrule
\end{tabular}
\end{table}

Note two things about the criteria in Table~\ref{tab:good-decision-support}. First, a \dst helps a decision maker. It does not necessarily provide the answers for any of the criteria. Second, it goes beyond recommendations and judgements, which are typically the focus in XAI; largely because these are considered to be some of the harder problems. 

\subsection{Cognitive processes for decision making}

\citet{Hoffman2022-en} argue that, when a person is trying to determine why the system produced a particular output, they engage in the cognitive process called \textbf{abductive reasoning}. Abductive reasoning is the process of forming hypotheses and judging their likelihood for the purpose of explaining observations or facts \citep{Peirce2009-zg}. This abductive process is engaged as soon as people start interacting with a system, irrelevant of any particular explainability or interpretability tools. As such, \citeauthor{Hoffman2022-en} argue that a model of abductive reasoning is a suitable foundational model for conceptualising explainable AI.

\begin{figure}[!th]
\centering
\includegraphics[scale=0.5]{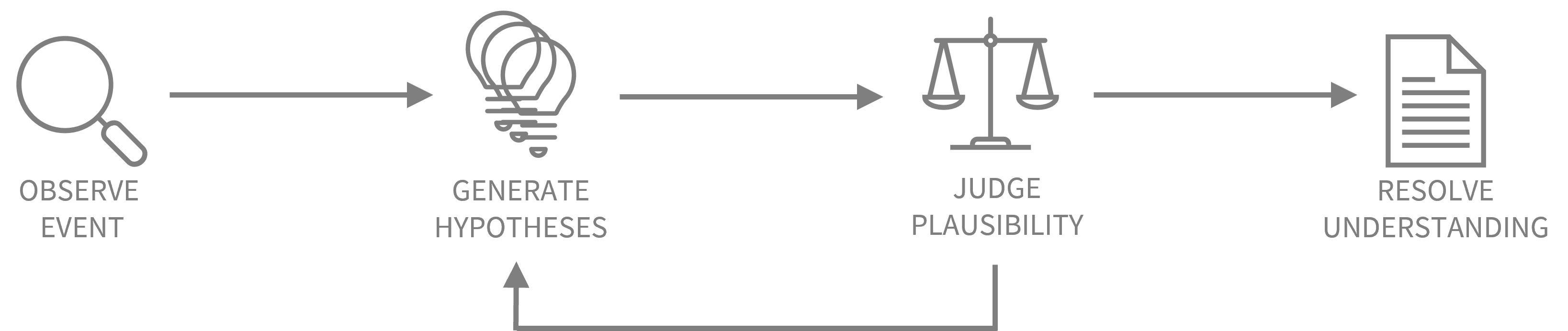}
\caption{A model of abductive reasoning. When someone observes an event  they do not understand, they generate hypotheses of the cause, and then judge the plausibility of (some of) these hypotheses; potentially iterating to generate more hypotheses as new evidence is found. Eventually, the mis-understanding is resolved, possibly to be re-visited later.}
\label{fig:abductive-reasoning}
\end{figure}

\citet{Peirce2009-zg} defines the process of \textbf{abductive reasoning} as five step process, outlined in Figure~\ref{fig:abductive-reasoning}\footnote{The final step of \emph{extending} the explanation is omitted because it simply repeats the process when new events are observed.}:
\begin{enumerate}
    \item \emph{Observe an event or phenomenon}: A person observes an event, usually one that is surprising and does not fit their current mental model. This leads to them to start searching for  explanatory hypotheses.
    \item \emph{Generate hypotheses}: The person generates some potential reasons for why they would have observed the event/phenomenon, which are at this point, informed guesses. The reasons are \emph{hypotheses}.
    \item \emph{Judge the plausibility of the hypotheses}: The person searches for evidence that may support or undermine different hypotheses, perhaps ruling out some or making some more likely.  Some hypotheses are ruled out, while others become are judged as more likely. This process may lead to new hypotheses.
    \item \emph{Resolve understanding}: A particular hypothesis fits the new observation and previous experience of the decision maker, so is adopted as the most likely cause of the observation. This may be a tentative resolution.
    \item \emph{Extend}: Revise and extend the process when new evidence is observed or considered.
\end{enumerate}

This philosophical model of decision making is supported by research from cognitive science. \citet{Klein2007-bs} present a theory of sensemaking known as the \emph{Data/Frame Theory}, built from studies with expert decision makers. The \emph{frame}, a generalisation of a hypothesis, is a model of how something works, while the \emph{data} contains observations made and inferences that combine the observations and the frame. The data is used to adjust the frame (similar to judging plausibility) and the frame is used to determine what new data to find. \citeauthor{Klein2007-bs} shows that people make decisions by first using their intuition to narrow down to a set of likely options (the frames), and then go through each option one-by-one, searching for evidence (data) to make judgements. \emph{Good} decision makers search for evidence that both supports and refutes a hypothesis. 
\citet{Klein2007-bs} argue that abductive reasoning plays a central role in this sensemaking process. We build the \newxai framework around the Data/Frame model.

\citet{Hoffman2022-en} argue that abductive reasoning is a suitable foundation  for conceptualising explainable AI.
In this paper, we argue for a similar --- yet orthogonal --- view; specifically, explainable AI for \textbf{decision making}. In decision making, the \textbf{decision maker is involved in two (related) reasoning processes}. First, the decision maker is trying to make a judgement/decision about the world (e.g.\ a diagnosis). The observed event is something that needs explaining, such as a medical symptom, the hypotheses are the potential causes of that observation, such as a disease, and the judgement is about the likelihood of a particular hypotheses being true given the evidence.
Second, the decision maker is also making a judgement about the \dst (e.g.\ whether its reasoning is sound, as proposed by \citet{Hoffman2022-en}). The observed event is an output $o$ of an AI system $f$ given an input $i$. The hypotheses characterise which combinations of inputs $i$ and computations (parts of $f$) caused this output to occur. The process of judgement is to determine why the system produced the output that it did. 
%
Table~\ref{tab:difference-between-xai-and-newxai} outlines this for a medical diagnosis scenario. Decision makers need to `invert' the explainable AI process to align with the decision-making process. \citeauthor{Hoffman2022-en} argue that machines reasoning in an abductive manner align better with this process.
The argument in this paper is similar, but rather than giving abductive explanations, \dsts should be designed to explicitly support the abductive reasoning process.

\begin{table}[!ht]
\centering
\caption{The different ways that Explainable AI and \Newxai align against the human abductive reasoning process, using medical diagnosis as an example.}
\label{tab:difference-between-xai-and-newxai}
\begin{tabular}{lll}
\toprule
\textbf{Human Reasoning Step} & \textbf{Explainable AI} & \textbf{\Newxai}\\
\midrule
Event to explain & AI diagnosis                   & Medical symptoms\\
Hypotheses       & AI reasoning                   & Medical conditions\\
Evidence         & Models, explanatory information & Evidence for/against hypotheses\\
Judgement        & Causes of inputs to AI diagnosis & Likelihood of medical conditions\\
\bottomrule
\end{tabular}
\end{table}

\subsection{Explainable/interpretable AI and decision making}
\label{sec:background:xai-and-decision-making}

As noted earlier, the initial assumption of AI-assisted \dsts is that, if we provide people with recommendations, they will follow those recommendations when required, leading to better decisions. However, issues such as warranted and unwarranted distrust \citep{Jacovi2021-nw} mean that AI systems are often deployed and then largely ignored \citep{Sivaraman2023-ea,Gunning2019-kj}. Explainable/interpretable AI helps to mitigate the issues of unwarranted distrust by helping people to understand why decisions are made, potentially building trust and improving decision making.
\citep{Miller2019-jw,Gunning2019-kj,Mueller2019-xg,Swartout1993-hk,Miller2022-ry}.

However, recent research has shown that the recommendation-driven explainable AI and interpretable machine learning has little effect on decision making \citep{Gajos2022-is,Bucinca2021-lq,Green2019-ic,Green2019-jk,Poursabzi-Sangdeh2021-yy,Nourani2021-al}, although this is not always the case \citep{Leichtmann2022-te,Madumal2020-sj,Van_der_Waa2021-yu}. 
The two primary issues are \emph{over-reliance} and \emph{under-reliance}. When  decision makers over-rely a machine, they accept its recommendations even in cases where they are wrong, caused by unwarranted trust \citep{Jacovi2021-nw}. This is often attributed to \emph{automation bias}, where the machine is considered correct because it is a machine, so ``must be right''. When decision makers under-rely a machine, they reject its recommendations even in cases where the recommendations are correct, caused by unwarranted distrust \citep{Jacovi2021-nw}. This is often attributed to \emph{algorithmic aversion}, where machine outputs are rejected but would be accepted had they come from a human. In either case, a poor decision can often be attributed to \emph{fixation} \citep{Klein2007-bs} on a particular hypothesis --- searching for evidence to support one hypothesis without considering others.

Some experimental studies \citep{Chromik2021-nj,Eiband2019-pf,Ghai2021-xw,Vera_Liao2021-fw} indicate that most participants  do not cognitively engage with explainability tools. Participants who do engage seem to believe that they understand explanatory information, leading to over-confidence in their decisions \citep{Chromik2021-nj}, even when presented with explanations that contain no useful information \citep{Eiband2019-pf}. 

We have seen this in observational studies. In a study where expert board gamers used an AI-based assistant to provide recommendations, we saw that many players did not truly engage with recommendations, and adding explanatory information did not help, as they typically did not engage with explanatory information either. When they did, we saw some behavioural changes, but not enough to see a statistically significant improvement in decision making.

Similarly, in experiments on a general task with general participants, \citet{Bucinca2021-lq} and \citet{Gajos2022-is} show that explainability did not mitigate over-reliance, and in some cases it increased over-reliance. They assert that this is because participants did not pay attention to explainability information. They propose three \emph{cognitive forcing} strategies to mitigate over-reliance, such as forcing people to give a decision before seeing a recommendation.
%
Their results showed that cognitive forcing slightly mitigated over-reliance, compared to feature-based and uncertainty judgements, particularly disregarding incorrect AI recommendations, but with no statistically significant differences in task performance. 
Interestingly, despite cognitive forcing functions being more effective, they were least preferred by participants. \citet{Bucinca2021-lq} attribute this to the phenomenon of people not wanting to exert mental energy \citep{Kool2018-tf}. This indicates that \Newxai could prove useful by not fixating people on particular recommendations, but allowing people to assess whether evidence supports their hypotheses, rather than trying to understand the \dst's reasoning.

These studies show that the recommendation-driven XAI does little to mitigate over- and under-reliance.
This has lead to several authors to argue that XAI research should return to the foundation of cognitive and social processes involved in decision making \citep{Kaur2022-cv,Vera_Liao2021-fw,Green2019-ic,Green2019-jk,Wang2019-xi}. 
Some authors \citep{Gajos2022-is,Bucinca2021-lq,Vera_Liao2021-fw} have attributed the cause of ignoring explanatory information to the `failure' of system 1 processing in dual process theory \citep{Kahneman2011-ra}. In brief, the theory is that system 1 thinking (fast, heuristic, and biased) `interferes' with system 2 thinking (slow and more accurate). The conclusion that is often, but not always drawn, is that \dsts should aim to support system 2 thinking and prevent system 1 `interference'.

However, \textbf{we caution against the idea of prioritising system 2 thinking over system 1 thinking}, for three reasons. First, research in \textbf{Naturalistic Decision Making} (NDM) demonstrates that `system 1' thinking, which they call \textbf{intuition}, is a powerful source of problem solving that often produces better and/or faster decisions than system 2 thinking \citep{Klein2015-pa}. For example, \citet{Coderre2003-vk} found that intuition significantly outperformed hypothetico-deductive reasoning in gastroenterology clinical diagnoses. \citet{Klein2017-uw} argues that experts make decisions by first using their intuitive problem solving ability (system 1) to generate likely options, and then use both system 1 and system 2 to assess whether their options are going to work. As such, any technique that aims to `override' intuitive thinking to prioritise structured thinking risks losing the benefits of expertise and prior knowledge.

Second, a recent article from \citet{Schulze2021-kg} argues that many studies that identify biases of system 1 thinking, such as base-rate fallacies, are partly a product of the predominant mode of experimental setup: participants receiving descriptive scenarios and being asked to give one-off judgements. \citeauthor{Schulze2021-kg} show that, in experiments where participants can practice a task, are involved in interactive problem solving with feedback, and can sample information, many cognitive biases are not as prevalent. Given that many AI-assisted decision support tasks would fall into the latter, the task and how cognitive biases affect decision making is important.

Finally, research demonstrating a lack of engagement with explanatory information runs experiments with general participants, not expert decision makers, and the consequences of poor decisions have low stakes. The effect of system 1 thinking may differ between contexts such as the experience of the decision maker and the stakes of the decision.

In short, we argue that we should not simply treat intuition as `bad' and prioritise structured thinking. \dsts should exploit the strengths of intuition and expertise, as well as structured thinking.

\section{(Explainable) AI as decision support}
\label{sec:xai-as-decision-support}

In this section, we provide a narrative of how we have arrived at the current status quo for decision support using explainable AI. We evaluate four difference approaches against the criteria of decision support identified in Table~\ref{tab:good-decision-support}. 
The four paradigms we discuss are: (1) giving recommendations with no explanatory information;  (2) giving recommendations with explanatory information;  (3) `intrinsically interpretable` models; and (4) cognitive forcing.

\subsection{Giving recommendations with no explanatory information}

The idea of decision support using AI came from the use of AI as automation. Given situations in which an AI model provides more accurate judgements/decisions than human decision makers, a simple of model of AI-assisted \dsts is to give recommendations, which decision makers can then take into account. Recommendations can be ranked lists of options such as in many recommender systems, but the typical model is to propose just the one recommendation that the AI model determines is the most likely or `best'. The assumption here is that for situations in which \dsts can handle many more factors simultaneously, it can provide insight  when the assumptions of the underlying modelling theory hold. This model is depicted below in Figure~\ref{fig:recommendation}.

\begin{figure}[H]
\centering
\includegraphics[scale=0.5]{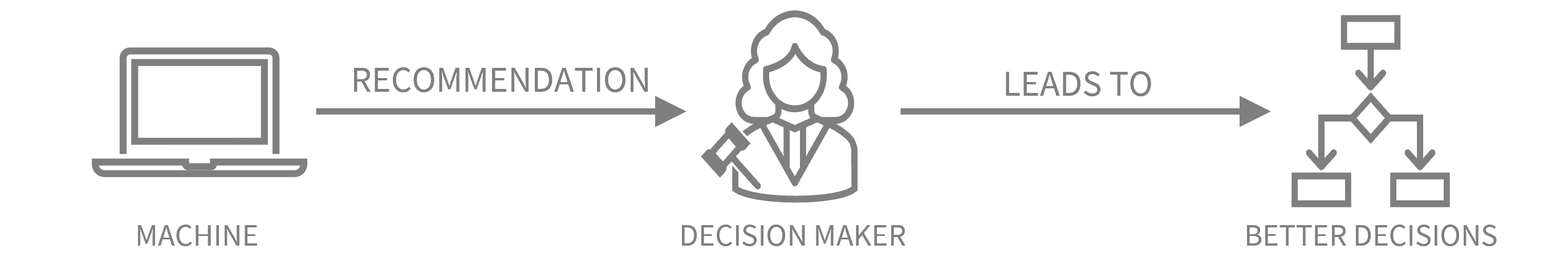}
\caption{A model of giving recommendations for decision support. This assumes that decision makers will carefully consider recommendations. However, empirical evidence suggests this is not the case.}
\label{fig:recommendation}
\end{figure}

The primary issue with this model was that, even if the \dst gives better judgements/decisions than the human, people tend to ignore this due to unwarranted distrust \citep{Miller2019-jw,Gunning2019-kj,Swartout1993-hk,Jacovi2021-nw,Hoff2015-ja,Hoffman2017-fn,Sivaraman2023-ea} or accept wrong decisions due to unwarranted trust \citep{Gajos2022-is,Bucinca2021-lq,Parasuraman1997-th}. The inability to scrutinise why decisions are made mean that the only information to rely upon to judge the correctness is: (1) extrinsic information such as accuracy measures; and (2) the decision makers' own expertise and knowledge. Given the former does not change from decision to decision, it is unsurprising that experts rely on their own expertise and novices over-rely on the tool \citep{Sivaraman2023-ea}.

Comparing this to the properties that comprise good decision support outlined in Table~\ref{tab:good-decision-support}, we can see why this approach is largely inneffective. Giving only recommendations:
\begin{enumerate}
    \item \textbf{does not} help to provide new options or to filter out unlikely options;
    \item \textbf{does not} consider possibilities or judgements beyond one option;
    \item \textbf{does not} help determine stakeholder values; 
    \item \textbf{does not} support making trade-offs; and 
    \item \textbf{does not} provide understanding of the machine decision.
\end{enumerate}

Can this still be useful? Of course! If a recommendation matches or is similar to our own judgement, it gives us confidence in our own decision, although this confidence may be misplaced. Further, if a recommendation is not similar to our own judgement, it may decrease our confidence and make us reconsider, potentially helping us to make a better decision. However, in either case, we receive no further support to help with the decision.

\subsection{Giving recommendations with explanatory information}

\begin{figure}[H]
\centering
\includegraphics[scale=0.5]{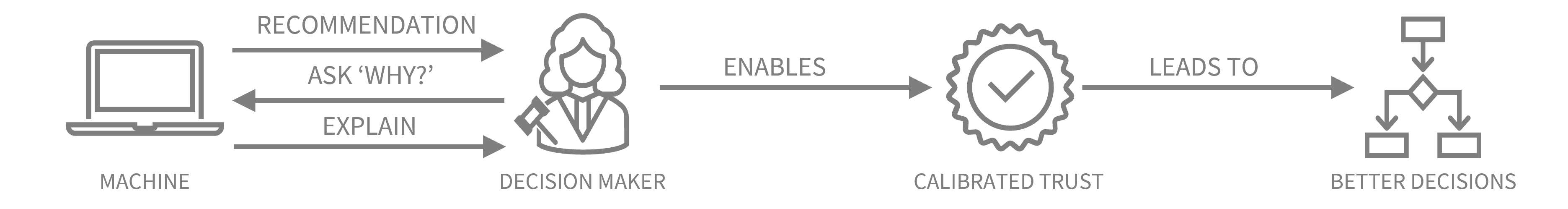}
\caption{A model of explainable AI for decision support. This assumes that the problem of distrust can be mitigates by giving reasons or explanations for decisions. However, empirical evidence suggests people do not pay careful attention to the reasons/explanations.}
\label{fig:recommendation-with-explainabilty}
\end{figure}

One solution to mis-calibrated trust is to provide explanatory information. This means giving justifications for decisions, providing evidence to support the decisions, and making models simple and easy to understand (see more on this in Section~\ref{sec:paradigms:interpretable-models}). A model of interaction between decision maker and \dst is shown in Figure~\ref{fig:recommendation-with-explainabilty}.

Comparing this to the properties that comprise good decision support outlined in Table~\ref{tab:good-decision-support}, we can see the criteria that the default model of XAI bring are that it:
\begin{enumerate}
    \item \textbf{does not} help to provide new options or to filter out unlikely options;
    \item \textbf{does not} consider possibilities or judgements beyond one option;
    \item \textbf{does not} help determine stakeholder values; 
    \item \textbf{partially} supports making trade-offs; and
    \item \textbf{does} provide understanding of the machine decision.
\end{enumerate}

So, compared to just recommendations, explainability provides understanding.
Contrastive explanations \citep{Miller2021-ix} partially support trade-offs when they answer `Why $A$ instead of $B$?', where $A$ is the output and $B$ a \emph{foil}. However, this only supports comparing each option the recommended one -- it is persuasive. It does not provide support for trading off other pairs of options. Later in Section~\ref{sec:newxai} we discuss how contrastive explanations could support trade-offs better.

Can this be useful? Of course! It  has the same properties as recommendations, but has the added benefit that if we agree with the recommendation, we can check that we agree for similar reasons; and if we disagree, we can check the reasons why, potentially helping us to improve our initial decision.

\subsection{Giving recommendations with an interpretable model}
\label{sec:paradigms:interpretable-models}

The third paradigm is to give recommendations using an `interpretable' model. \citet{Rudin2019-ee} argues that for high-stakes decisions, people should avoid using black-box models all together, and instead should use interpretable models. Interpretable models use a small set of features with a low complexity of interaction between them. 

The model of interaction with an interpretable model is same as for recommendations with explanatory information, outlined in Figure~\ref{fig:recommendation-with-explainabilty}. The assumption is that because the model is interpretable, it contains most of the explanatory information that is required.
However, it is difficult to claim that a model being interpretable is really \emph{decision support}. The model's options, possibilities, and trade-offs would have to be calculated by the decision maker. Assistance to calculate these is a combination of interpretability and explainability, so overlaps with the previous section.

Comparing this to the properties that comprise good decision support outlined in Table~\ref{tab:good-decision-support}, we can see the criteria that the interpretability brings are that it:
\begin{enumerate}
    \item \textbf{does not} help to provide new options or to filter out unlikely options;
    \item \textbf{does not} consider possibilities or judgements beyond one option;
    \item \textbf{does} provide understanding of the machine decision;
    \item \textbf{does not} help determine stakeholder values; and
    \item \textbf{does not} support making trade-offs.
\end{enumerate}

\subsection{Cognitive forcing}

The final paradigm is cognitive forcing. The assumption of cognitive forcing is  that forcing people to engage with the decision cognitively can mitigate over-reliance. \citet{Gajos2022-is} and \citet{Bucinca2021-lq} implement this idea using three approaches, as outlined in Section~\ref{sec:background:xai-and-decision-making}. The commonality between these three approaches is that decisions are initially withheld from the decision maker, but explanatory information is provided, forcing the decision maker  to engage with the explanatory information. This model is depicted below in Figure~\ref{fig:cognitive-forcing}. Withholding recommendations `forces' the decision maker to cognitively engage with the decision and therefore, to consider different options and make trade-offs. Giving explanatory information from the start may help the decision maker to focus on useful information.

\begin{figure}[H]
\centering
\includegraphics[scale=0.5]{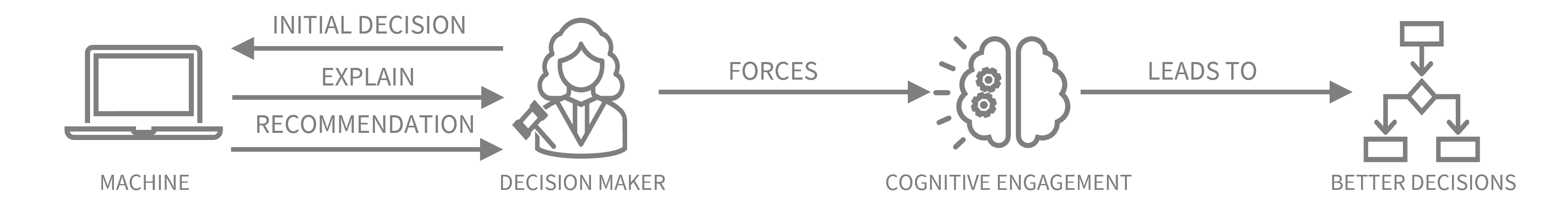}
\caption{A model of cognitive forcing for decision support. This assumes that by withholding the decision (for perhaps a short period) and optionally giving an `explanation', it forces people to engage, limiting over-reliance. However, current solutions are still recommendation-driven because they `explain' just the machine decision.}
\label{fig:cognitive-forcing}
\end{figure}

Comparing this to the properties that comprise good decision support outlined in Table~\ref{tab:good-decision-support}, we can see the criteria that cognitive forcing brings are that it:
\begin{enumerate}
    \item \textbf{partially} helps to provide new options or to filter out unlikely options by forcing the decision maker to do so;
    \item \textbf{does not} consider possibilities or judgements beyond one option;
    \item \textbf{does not} help determine stakeholder values;
    \item \textbf{partially} supports making trade-offs; and
    \item \textbf{does} provide understanding of the machine decision.
\end{enumerate}

While this is an improvement beyond the default XAI approaches, weaknesses remain. First, this approach is still an instance of our friend Bluster, rather than our friend Prudence -- it is just that Bluster pauses  before stating their opinion. Second, it does not provide helpful decision-making information for options other than the recommendation.

Next, we propose a model that builds on the strength of cognitive forcing, but on a \textbf{foundation of the cognitive science of decision making}.

\section{\Newxai: A conceptual framework of explainable decision support}
\label{sec:newxai}

In this section, we present a new conceptual framework of hypothesis-driven explainable decision support called \textbf{\newxai}. The two primary design criteria for this are:
\begin{enumerate}
    \item Support the properties of good decision making outlined in Section~\ref{sec:background:xai-and-decision-making}. That is, support the decision maker's \textbf{cognitive decision making process}.
    \item Provide better \textbf{internal locus of control} \citep{Shneiderman2016-bn} to the decision maker about which options to explore and when.
\end{enumerate}

\noindent
To do this, we propose that the vision of \newxai is to:

\begin{quote}
    \textbf{Support the decision maker to access the information they want and need to evaluate a hypothesis, when they want it.}
\end{quote}

\subsection{Conceptual framework}

\begin{figure}[H]
\centering
\includegraphics[scale=0.5]{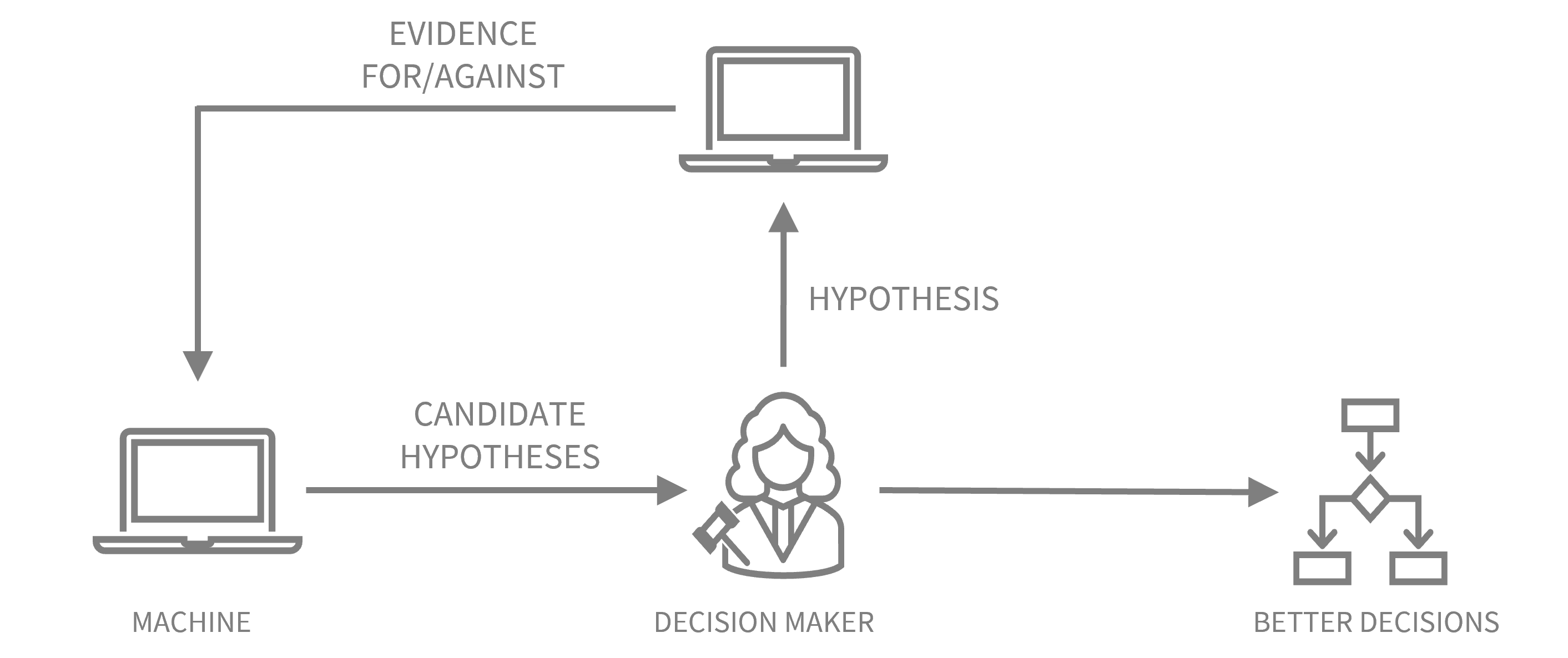}
\caption{A model of \Newxai. This aims to explicitly align with the decision making process, keeping the decision maker in control and aiming to have users rely on evidence instead of recommendations.}
\label{fig:evaluative-ai}
\end{figure}

The conceptual framework for \newxai is shown in Figure~\ref{fig:evaluative-ai}.  \Newxai does not (necessarily) provide recommendations, but instead offers support to filter out unlikely options, generate new hypotheses, or both. The decision maker then analyses a hypothesis, asking the \dst to provide evidence for and against the hypothesis.  Evidence could also be presented contrastively: what is the evidence for/against a particular hypothesis rather than some other hypothesis. Important,the decision maker to maintains \textbf{control} over which hypotheses to explore.

\subsubsection{Options}
Existing explainable AI approaches tend to provide one or sometimes two options to consider. The \newxai paradigm argues that the presentation of options should be context-specific. For example, given a probabilistic classifier, a \newxai system could highlight options that are within a certain probability of the most likely, perhaps withholding the probabilities themselves. Alternatively, it could use uncertainty estimates, which are known to help decision making \citep{Tomsett2020-kl}, and from that, determine which options to filter.

Approaches such as this help to reduce fixation because there is no single recommendation. Cognitive forcing takes a similar approach, but current cognitive forcing techniques  are recommendation-driven \citep{Gajos2022-is,Bucinca2021-lq}.

\subsubsection{Judgement support}
All five paradigms support judgements. The \newxai framework puts the human at the centre of the judgement process. In this framework, the judgement is made by the human decision maker with support from the \dst, which gives feedback (evidence for/against) of proposed hypotheses. 
This is \textbf{supporting judgement}, rather than \textbf{giving judgement}.
The difference is who owns control over which hypotheses to explore. In a recommendation-driven approach, both the human and machine provide a judgement. In the \newxai paradigm, the human provides judgement and the machine provides feedback.

\subsubsection{Trade-off support}
Contrastive explanation trades-off different outcomes. However, these explain only why non-recommended options are `incorrect'. This is a `persuasive' approach --- it justifies why the machine's decision is correct. The \newxai framework instead:
\begin{enumerate}
    \item explains trade-offs between any two sets of options; and
    \item provides evidence both for \emph{and} against each option, irrelevant of the judged likelihood of that option.
\end{enumerate}

Good decision makers assess an option by looking for evidence that supports it, but also evidence that refutes it. For example, in a study with anaesthesiology residents, \citet{Rudolph2003-tu} showed that participants who fixated on an initial option did poorly, but that participants who kept an open mind on all options do not perform much better. Instead, those who jump to an initial conclusion and test it, looking for \emph{negative evidence}, make the best decisions. Recommendation-driven approaches typically do not provide evidence against the recommendation, nor evidence supporting other options. \Newxai supports both, rather than aiming to persuade that the machine is correct.

Supporting trade-offs has gained interested in recent years under the name \textbf{option awareness}. Similar to the better-known \emph{situation awareness} \citep{Endsley2017-tq}, option awareness is the analysis and understanding of various options and their relative trade-offs \citep{Pfaff2013-an}. \citet{Pfaff2013-an} show that using visual analytics to allow decision makers to explore options increases their option awareness, resulting in more accurate and faster decisions. More recently, \citet{Drury2022-ic} proposed a framework in which machines consider their own option awareness to support human decision makers. However, research into machine-assisted option awareness is still in its infancy.

\subsection{Example: Diagnosis}
\label{sec:new-xai:example}

Consider an example of a \dst for diagnosing skin cancers, such as the ISIC 2018 lesion diagnosis challenge\footnote{See \url{https://challenge.isic-archive.com/landing/2018/47/}.}. There are seven possible disease categories: 
(1) melanoma; 
(2) melanocytic nevus;
(3) basal cell carcinoma;
(4) actinic keratosis;
(5) benign keratosis;
(6) dermatofibroma; and
(7) vascular lesion.
Given a dermoscopic image of a legion along with some meta information, such as which area of the body the lesion is found, the task is to diagnose the most likely category.

\begin{figure}[!th]
\centering
\includegraphics[scale=0.45]{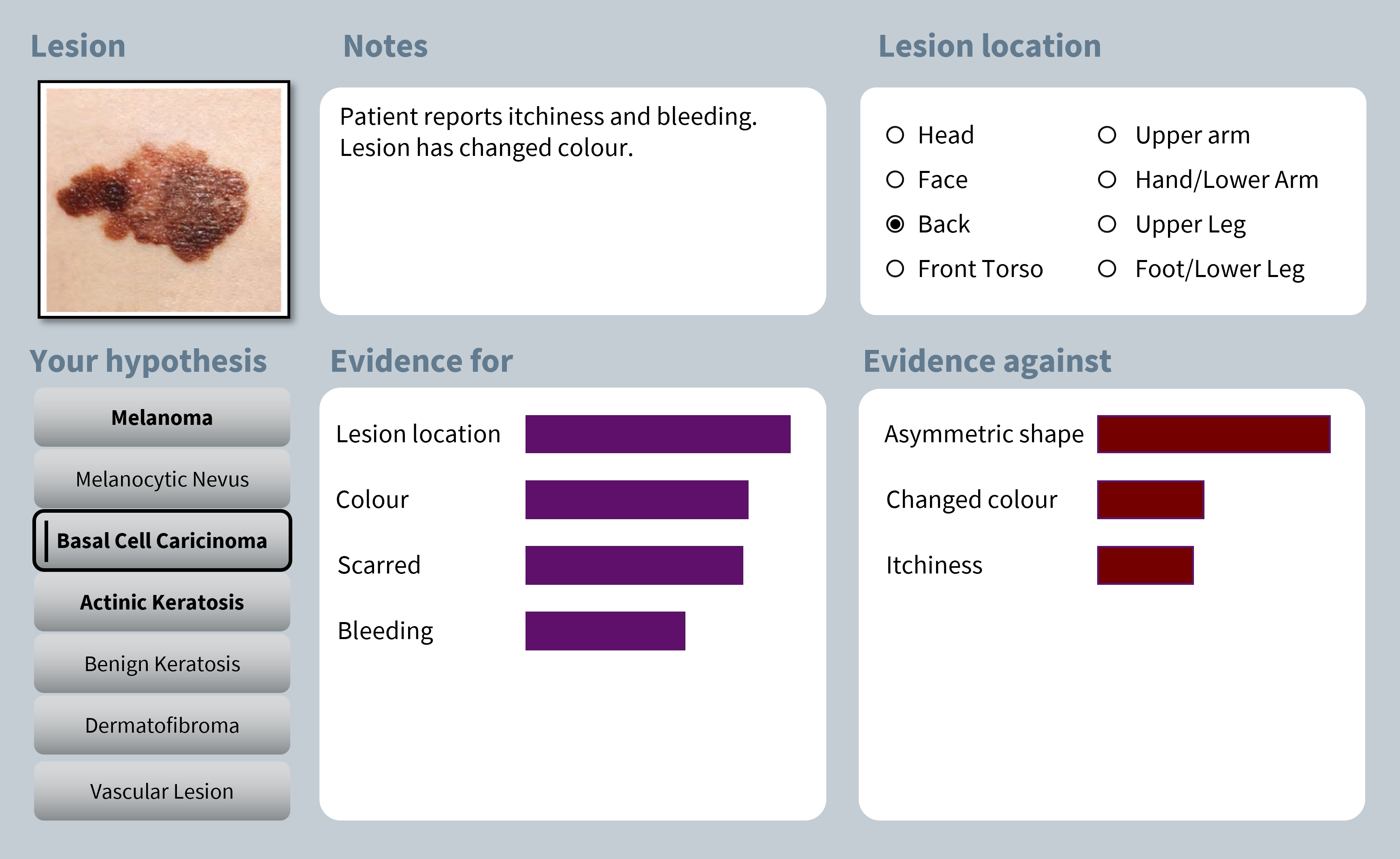}
\caption{A simple prototype of a diagnostic interface using \newxai}
\label{fig:interface}
\end{figure}

Figure~\ref{fig:interface} shows a simple prototype interface for such a system. 
In this case, there are seven potential hypotheses/diagnoses. The \dst filters out those that are unlikely, leaving just three: melanoma, basal cell carcinoma (BCC), or actinic keratsosis (AK)\footnote{It is unlikely that these three would be the case as AK is quite different from BCC and melanoma, but this makes for a good illustrative example.}, highlighted with bold text. By highlighting only likely hypothesis instead of just one hypothesis, this could help to mitigate fixation on just one option.

Hypotheses can be explored. In Figure~\ref{fig:interface}, the location of the lesion (back), its colour, scarring, and that it sometimes bleeds, is strong support for a BCC diagnosis. However, the asymmetric shape, itchiness, and recent colour change are evidence against BCC. 
Other forms of evidence could be provided to the decision maker, such as finding similar instances using case-based reasoning \citep{Wang2019-xi,Sormo2005-vq}.
It is up to the decision maker to make the final decision, integrating their expert knowledge and the information/evidence from the \dst.

\subsection{Summary} 

Comparing model to the properties that comprise good decision support outlined in Table~\ref{tab:good-decision-support}, we can see the criteria that \newxai brings are that it:
\begin{enumerate}
    \item \textbf{does} help to provide new options or to filter out unlikely options by forcing the decision maker to do so;
    \item \textbf{does} help to identify possibilities and support judgement for options;
    \item \textbf{does not} help determine stakeholder values; 
    \item \textbf{does} support making trade-offs; and
    \item \textbf{does} provide understanding of the machine decision.
\end{enumerate}

\begin{table}[!t]
\centering
\caption{A summary of the decision support provided by different paradigms. Partial means that this provides information for the recommended option only. \Newxai explicitly provides support to explore options and perform trade-offs. XAI and cognitive forcing allow this if they support contrastive explanation. In \newxai, it should be default.}
\label{tab:comparison}
\begin{tabular}{lcccccc}
\toprule
  &\multicolumn{6}{c}{\textbf{Support provided}}\\
 \cmidrule{2-7}
 \textbf{Decision support} & \textbf{Options} & \textbf{Possibilities} & \textbf{Judgement} &\textbf{Value}  & \textbf{Trade-offs} & \textbf{Understandable}\\
\midrule
Recommendation     & \no  & \yn  & \yn  & \no  & \no  & \no  \\
XAI                & \no  & \yn  & \yn  & \no  & \yn  & \yes \\
Interpretability   & \no  & \yn  & \yn  & \no  & \no  & \yes \\
Cognitive forcing  & \yn  & \yn  & \yn  & \no  & \yn  & \yes \\
\Newxai            & \yes & \yes & \yes & \no  & \yes & \yes \\
\bottomrule
\end{tabular}
\end{table}

Table~\ref{tab:comparison} compares the five approaches to explainable \dsts. \Newxai explicitly aims to provide support for options, judgement, understanding, and trade-offs.
The difference between cognitive forcing and \newxai is two-fold:
\begin{enumerate}
   \item \textbf{Cognitive processes}: \Newxai is built on the Data/Frame model of sensemaking \citep{Klein2007-bs}, so supports the decision maker's cognitive process by allowing them to explore hypotheses, rather than providing only the information to the justifies a machine recommendation. 
    \item \textbf{Control}: \Newxai explicitly hands control of which hypotheses are investigated and prioritised to the decision maker, resulting a \textbf{machine-in-the-loop} paradigm \citep{Green2019-ic,Green2019-jk}, rather than a human-in-the-loop paradigm. 
\end{enumerate}

For this reason, we assert that \newxai obtains the benefits of cognitive forcing, while also giving control to the decision maker to explore the strengths and weaknesses of any option, rather than just the strengths of the recommendation and the weaknesses of alternatives.

\subsection{Long live explainable AI!}

This article proposes a pivot from recommendation-driven decision support to hypothesis-driven decision support. Does this imply that explainable AI is dead? We do not believe so.

First, there are applications of XAI beyond decision support, such as verification/regulation, scientific discovery, generating insights about underlying data, etc. Further, there are applications where recommendation-driven approaches may be the best way to improve decisions, such as making decisions at scale.

Second, for a machine to judge decisions, we will need an underlying  decision-making model, meaning that existing AI techniques such as machine learning, planning, and optimisation will be required, along with their XAI techniques.

Third, \newxai \textbf{is} explainable AI. A lot of cognitive and social aspects of good explainability apply to hypothesis-driven decision support. For example, the four key properties of human-to-human explanation \citep{Miller2019-jw}: (1) explanations are contrastive; (2) explanations are selected; (3) explanations are interactive; and (4) explanations are causal --- all still apply in providing evidence and evaluating trade-offs.

Fourth, from the earlier example, it is clear that many existing tools will play a part in \newxai: constrastive explanation \citep{Miller2021-ix}, Weights of Evidence (WoE)  \citep{Melis2021-kt},  feature importance, case-based reasoning techniques \citep{Sormo2005-vq}, etc., all  generate evidence. New models of XAI are needed, but existing work lays a solid foundation.

\subsection{Challenges and Limitations}

There are clear potential limitations with this approach. First, if people tend to dismiss recommendations and any explainability information, why would they pay attention to evidence? The \newxai framework makes the same assumptions as earlier work: that people care at all about what a machine has to say. A very real risk is that decision makers will not engage with a tool that supports cognitive reasoning either. However, one argument against this is that this \newxai provides decision makers with better \textbf{control} \citep{Vered2020-cv}, and follows a process that they will \textbf{naturally follow}, rather than recommendation-driven approaches, which somewhat disrupt the decision-making process.

Second, it is difficult to imagine \newxai solutions  will result in lower cognitive load. A strength of the recommendation-driven approaches is that they reduce the information that a decision maker needs to consider to just the most relevant. \Newxai will likely result in designs that force more engagement with the decision, but less preferred by decision makers \citep{Bucinca2021-lq}. However, following a model of abductive reasoning can still reduce information compared to having no decision support; e.g. presenting only the most likely hypotheses; prioritising access to the most important information, etc. Striking this balance is a challenge.

\subsection{An incomplete research agenda}

In this section, we present an incomplete research agenda based around the \newxai framework.

\subsubsection{Observing events}

The first step in abductive reasoning is to observe an event. While it is difficult to control the attention that a decision maker pays to data/events, we can support this in several ways:

\begin{enumerate}
    \item \textbf{Design interfaces to make it clear what has happened, what data is being used}, etc.\ \citep{Hoffman2022-en}. This includes allow people to explore relevant attributes of a system.
    \item \textbf{Highlight anomalous behaviour/events}, as these are typically the events that people require to make decisions or to understand \citep{Miller2019-jw}.
\end{enumerate}

\subsubsection{Generating options}

The ability for the machine to put forward options can help decision makers in two ways. First, it can present options that decision makers did not consider. Second, it can speed up decision making in time-sensitive environments by filtering out some options and/or allowing options to be assessed more systematically.

Generating options can be supported in several ways beyond providing 1-2 recommendations or a ranking:

\begin{enumerate}

    \item \textbf {Provide probabilities over options:} This can help the decision maker to narrow down likely options, but is perhaps subject to over-fixation (on the most likely option) in the same way that recommendations are.

    \item \textbf{Provide a set of likely options:} This can help the decision maker to narrow down the set of hypothesis without giving a recommendation, as in the Example in Section~\ref{sec:new-xai:example}, reducing  fixation.
    
    \item \textbf{Provide uncertainty measures:} The \dst provides its uncertainty about its filtered options (the decision could be provided or not) to allow healthy scepticism of the model \citep{Zhang2020-uy}.

    \item \textbf{Intervention}: The \dst does not initially narrow down options, but allows the decision maker to explore and select their answer, and then intervenes if it disagrees above some threshold, prompting the decision maker to consider the \dst's most likely responses.

    \item \textbf{Relate inputs and hypotheses:} The decision maker selects a subset of inputs they believe are important and the tool shows which hypotheses are supported or denied by that evidence.

\end{enumerate}

\subsubsection{Judging plausibility}

Judging the plausibility of outcomes can be supported several ways by a \dst

\begin{enumerate}

 \item \textbf{Explainable/interpretable AI}: Provide reasoning steps that link the evidence to the hypothesis, such as rules, decision steps, or explanations.
 \item \textbf{Provide evidence weights}: Show how different inputs weight positive and negatives to support a decision.
 \item \textbf{Provide epistemic uncertainty:} Show uncertainty  in the form of e.g.\ an uncertainty measure using entropy.
 \item \textbf{Provide aleatoric uncertainty}: Show a measure of uncertainty of the evidence itself.
 \item \textbf{Evidence selection:} Allowing the decision maker to change evidence (inputs) and ask the \dst to re-evaluate.
 \item \textbf{Argumentation:} The decision maker identifies evidence that supports (refutes) a hypothesis, and the \dst highlights: (a) other evidence that strongly refutes (supports) that hypothesis; and (b) other hypotheses that are strongly supported (refuted) by that evidence.

\end{enumerate}

\subsubsection{Resolution and re-evaluation}

For any \dst, it is important that justification for that decision is recorded. Allowing the decision maker to record: (a) the outcome; (b) evidence for/against that outcomes; and  (c) alternatives that were not chosen and why; is important for both decision resolution and also re-evaluation. Note that recording evidence must include evidence that is used by the human decision maker but not available to the \dst. Further research is required to determine how to support decision makers when re-evaluating a decision to come up to speed quickly and effectively.

\section{Conclusion}
\label{sec:conc}

Our friends Bluster and Prudence can both be useful in supporting our decisions, but Prudence puts us in control and is better at helping us weigh up different options. Bluster can be useful; for example, in low stakes decisions or if time is limited; but overall, Bluster does not support our cognitive process as well as Prudence. This paper calls for AI-assisted \dsts to follow the lead of Prudence, and it presents a new conceptual  framework of machine-in-the-loop \dsts. We call this conceptual framework \newxai.

However, \newxai is not a panacea for \dsts. Our friend Prudence makes us do more work, which people prefer to avoid \citep{Kool2018-tf}. It is likely that experimental studies will reflect the study from \citet{Bucinca2021-lq} and show that participants opt for less work, but with worse results. This is a side effect of having people cognitively engage with decisions; however, one that we may need to accept if we want to truly improve AI-assisted decision support.

We conclude by repeating that the current paradigm of explainable AI as justified recommendations is `dead'. But the new paradigm that includes hypothesis-driven explainability could take the thrown, so long live explainable AI!

\bibliographystyle{unsrtnat}
\bibliography{bib}

\end{document}